\documentclass{article}

\usepackage[preprint]{neurips_2023}

\usepackage[utf8]{inputenc} 
\usepackage[T1]{fontenc}    
\usepackage{hyperref}       
\usepackage{url}            
\usepackage{booktabs}       
\usepackage{amsfonts}       
\usepackage{nicefrac}       
\usepackage{microtype}      
\usepackage{xcolor}         
\usepackage{graphicx}
\usepackage{animate}
\usepackage{float}
\usepackage{amsmath}
\usepackage{amssymb}
\usepackage[ruled,vlined]{algorithm2e}
\usepackage{multirow}
\usepackage{colortbl}
\usepackage{array}
\usepackage{makecell}
\usepackage{bbding}
\usepackage{enumitem}
\usepackage{caption}
\usepackage{tabularx}
\usepackage{subfig}

\def\loss{{\mathcal{L}}}

\newcommand{\Tref}[1]{Tab.~\ref{#1}}
\newcommand{\Eref}[1]{Eq.~(\ref{#1})}
\newcommand{\Fref}[1]{Fig.~\ref{#1}}

\def\onedot{.}

\def\etc{\emph{etc}\onedot}

\newcommand{\methodnamenocolor}{EvPlug}

\newcommand{\methodname}[1]{\textbf{EvPlug}}

\title{\methodnamenocolor{}: Learn a Plug-and-Play Module for Event and Image Fusion}

\author{%
  Jianping Jiang$^{1,2}$, Xinyu Zhou$^{3}$, Peiqi Duan$^{1, 2}$, Boxin Shi$^{1, 2}$ \\
    $^{1}$National Key Laboratory for Multimedia Information Processing, School of Computer Science\\
    $^{2}$National Engineering Research Center of Visual Technology, School of Computer Science\\
    $^{3}$National Key Lab of General AI, School of Intelligence Science and Technology\\
     Peking University\\
    \texttt{alanjjp98@gmail.com}, \texttt{ \{zhouxiny, duanqi0001, shiboxin\}@pku.edu.cn }\\
}

\begin{document}

\maketitle

\begin{abstract}
 Event cameras and RGB cameras exhibit complementary characteristics in imaging: the former possesses high dynamic range (HDR) and high temporal resolution, while the latter provides rich texture and color information. This makes the integration of event cameras into middle- and high-level RGB-based vision tasks highly promising. However, challenges arise in multi-modal fusion, data annotation, and model architecture design. In this paper, we propose \methodnamenocolor{}, which learns a plug-and-play event and image fusion module from the supervision of the existing RGB-based model. 
The learned fusion module integrates event streams with image features in the form of a plug-in, endowing the RGB-based model to be robust to HDR and fast motion scenes while enabling high temporal resolution inference.
Our method only requires unlabeled event-image pairs (no pixel-wise alignment required) and does not alter the structure or weights of the RGB-based model. 
 We demonstrate the superiority of \methodnamenocolor{} in several vision tasks such as object detection, semantic segmentation, and 3D hand pose estimation.
\end{abstract}

\section{Introduction}
\label{introduction}

\begin{figure}[t]
  \centering
    \begin{minipage}[t]{0.38\textwidth}
        \centering
        \includegraphics[width=\textwidth]{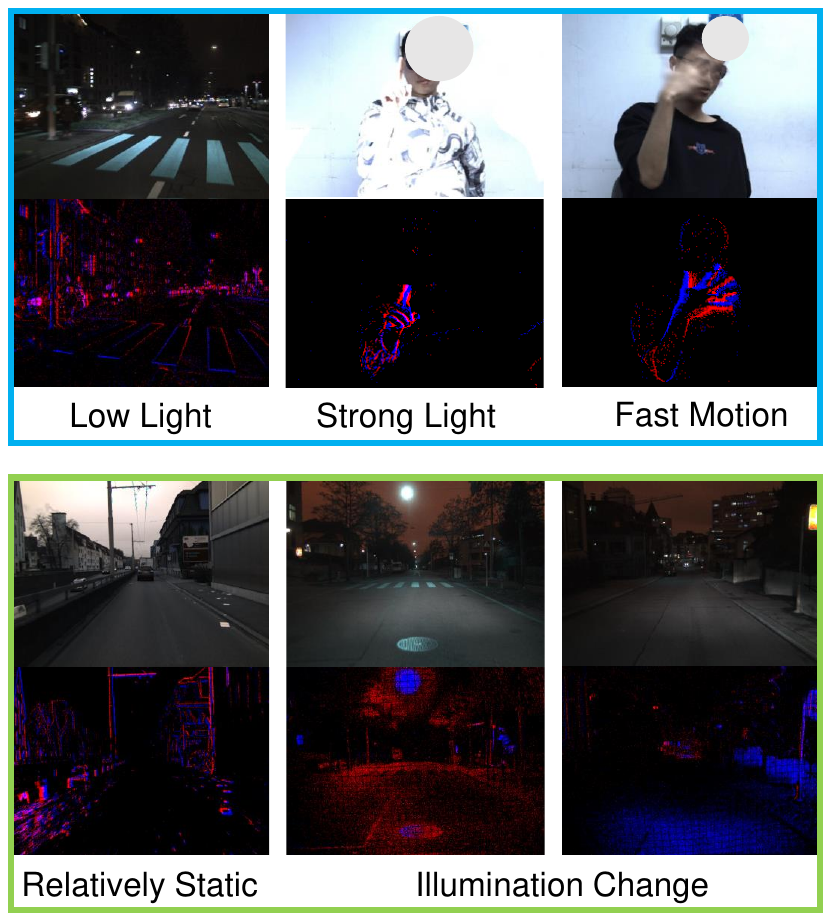}
        \caption{Complementary characteristics of images and events. Event cameras record more meaningful signals than RGB cameras in scenes shown in blue boxes, while results in green boxes are the opposite.}
        \label{fig: teaser complement}
    \end{minipage}
    \hfill
    \begin{minipage}[t]{0.58\textwidth}
        \centering
        \includegraphics[width=\textwidth]{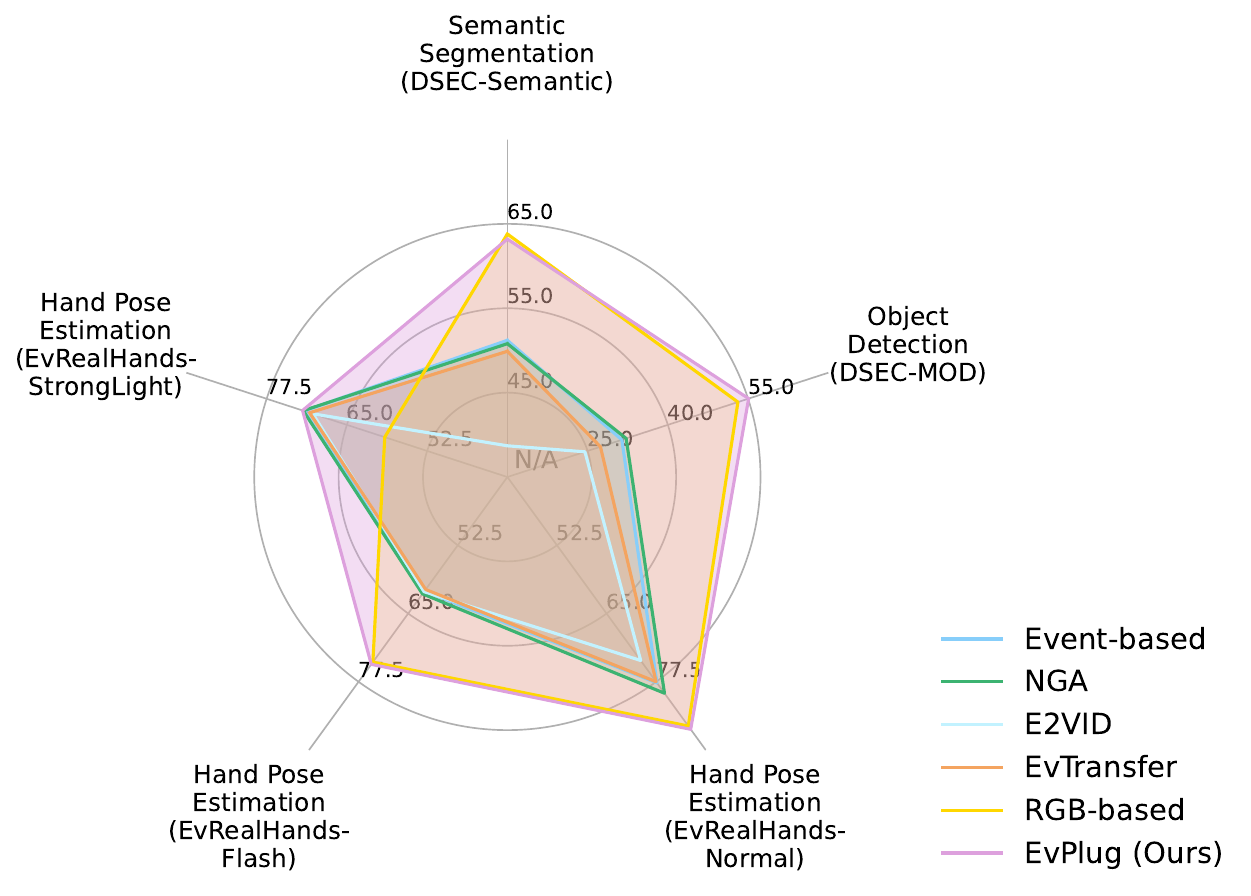}
        \caption{Radar plots of experimental results of \methodname{}, NGA~\cite{hu20nga}, E2VID~\cite{rebecq21e2vid}, EvTransfer~\cite{messikommer22evtransfer}, RGB-based method, event-based method on object detection (AP50 on DSEC-MOD~\cite{zhou2023rgb}), semantic segmentation (mIoU on DSEC-Semantic~\cite{sun22ess}), hand pose estimation (AUC on EvRealHands~\cite{jiang23evhandpose}).}
        \label{fig: radar compare}
    \end{minipage}
  \label{teaser}
  \vspace{-0.5cm}
\end{figure}

Traditional frame-based RGB cameras, featuring rich color and texture information as well as lower noise, are the mainstream sensors in computer vision research. However, due to their imaging mechanisms, they inevitably face issues such as overexposure, motion blur, and limited temporal resolution~\cite{han21evintsr, wang20jointfilter, hu20nga, Timelens, EDI}. Bio-inspired neuromorphic event cameras, with their asynchronous differential imaging mechanism (events are triggered by observing pixel-wise brightness changes exceeding a threshold in logarithmic domain), possess high dynamic range (HDR), high temporal resolution, low data redundancy, and low power consumption~\cite{licht08davis, dvs640, davis346, survey}. However, the imaging quality of event cameras degrades when the camera and scene are relatively static or when the scene illumination changes significantly~\cite{rebecq21e2vid, e2sri-cvpr20, Timelens}. These two types of cameras exhibit complementary characteristics in imaging as shown in \Fref{fig: teaser complement}, which has been demonstrated in various vision tasks, such as high frame-rate video reconstruction~\cite{Timelens, tulyakov22timelens++, gef-tpami}, HDR image restoration~\cite{han2020hdr}, object detection~\cite{zhou2023rgb}, depth estimation~\cite{daniel21ramnet}, \etc


However, event cameras have been studied for a much shorter period of time than RGB cameras, limiting the progress of their integration with RGB cameras.
Researchers have built large-scale datasets~\cite{deng09imagenet, lin14coco} and various meticulously designed network architectures~\cite{he16resnet, vaswani17transformer} based on RGB images, while the corresponding tasks in event-based vision are at a much less mature stage. Therefore, when fusing the two modalities of data, challenges arise from two perspectives: 1) data perspective: introducing event cameras requires collecting new data sequences and annotation, and large-scale data annotation implies high costs; 2) model perspective: designing new fusion algorithms and retraining models for event streams require additional model design works and computational resource.

To alleviate the challenges, existing methods~\cite{hu20nga, wang21dual, wang21evdistill, messikommer22evtransfer, gehrig20v2e, rebecq21e2vid} utilize the concept of domain adaptation~\cite{messikommer22evtransfer} (also referred as transfer learning, knowledge distillation in event-based vision~\cite{wanglin22kd_survey}), leveraging the consistency of visual signals between event and RGB cameras in geometric and semantic dimensions to transfer RGB-based knowledge to event-based tasks. 
They have achieved promising results on middle- and high-level vision tasks such as classification~\cite{messikommer22evtransfer}, object detection~\cite{wang21evdistill, messikommer22evtransfer, hu20nga}, and semantic segmentation~\cite{sun22ess}, where the input data can be compressed into high-dimension features\footnote{Event and RGB image fusion for low-level vision tasks such as video interpolation~\cite{Timelens, tulyakov22timelens++}, debluring~\cite{teng22nest, sun22event-based}, often utilize physical constraints and require pixel-wised alignment, so they do not apply domain adaptation.}.
However, these methods are insufficient in two aspects: 1) difficulty in handling the differences between these two types of visual information -- event streams contain rich motion and temporal information but lack color and texture information, while RGB images are the opposite~\cite{messikommer22evtransfer, zhu21eventgan}; 2) inability to achieve modality fusion that complementarily utilizes both modalities.

To conquer the limitations of the aforementioned domain adaptation methods, we propose \methodname{}, a framework that learns a pluggable event and image fusion module to strengthen the capability of the existing RGB-based model with events under various challenging scenes. 
In terms of event-and-image connection, unlike previous approaches that utilize the visual similarity between RGB images and event streams, we use the event generation model to constrain their relationship. This not only physically aligns with the differential imaging mechanism of the event camera but also theoretically endows the existing RGB-based model with the same temporal resolution as the event stream with temporal consistency. 
In terms of modality fusion, we employ event features to calibrate RGB features in the feature dimension to assure event-image quality consistency. On the one hand, this allows the event information to correct the feature space distortion caused when RGB image degrades, achieving modality complementarity. On the other hand, as a feature-level fusion method, it does not change the structure and weights of the RGB-based model, making the learned events and image feature fusion module plug-and-play during evaluation. 

The above information constraints and feature fusion can be unified into a single framework. We compare our method with other domain adaptation methods and supervised RGB-based and event-based methods in several downstream tasks, and the experimental results confirm the effectiveness of our approach as shown in \Fref{fig: radar compare}.
In summary, our contributions are as follows:
\begin{itemize}   
    \item a unified framework that learns a plug-and-play module under the supervision of existing RGB-based models connected by the event generation model;
    \item a flexible and generalized event and image feature fusion strategy to retain both merits of event and RGB cameras; and
    \item an image and event distillation benchmark, which includes the results of six methods in three downstream vision tasks.
\end{itemize}

\section{Related Works}
\label{related works}

\subsection{RGB-Event Fusion Methods}
Due to the high dynamic range, high time resolution, low redundancy, and low power consumption, recent researches have shown the potential of event cameras in object detection~\cite{anton18detection, zhou2023rgb, prophesee}, semantic segmentation~\cite{alonso19evsegnet, gehrig21dsec, sun22ess}, depth estimation~\cite{daniel21ramnet}, optical flow estimation~\cite{zhu18evflownet, zhu19evflow}, denoising~\cite{wang20jointfilter}, motion segmentation~\cite{mitrokhin2020learning, motion-seg-iccv19}, object detection~\cite{prophesee}, frame interpolation~\cite{han21evintsr}, human/hand pose estimation\cite{zou21eventhpe, victor21eventhands, jiang23evhandpose}, \etc{} 
Despite these advantages, event cameras exhibit higher noise levels due to less mature fabrication processes compared with RGB cameras, and they struggle to capture effective information when the camera and scene are relatively static or when the scene's illumination undergoes significant changes. Consequently, some studies attempt to fuse RGB images with event streams to enhance the robustness of vision models. 
Researchers have investigated various pixel-level alignment and fusion methods for low-level vision tasks, such as video frame interpolation~\cite{tulyakov22timelens++}, deblurring~\cite{sun22event-based}, and HDR imaging~\cite{messikommer22multi-bracket}, \etc{} 
In middle- and high-level vision tasks, hierarchical feature-level fusion methods are conducted on object detection~\cite{tomy22fusing, zhou2023rgb}, depth estimation~\cite{daniel21ramnet}, \etc{}
EventCap~\cite{lan20eventcap} utilizes events to track estimated human poses from images.
However, these RGB and event fusion methods are supervised for a specific task given ground truth (GT) with limited generality.

\begin{figure}[t]
  \centering
  \includegraphics[width=\textwidth]{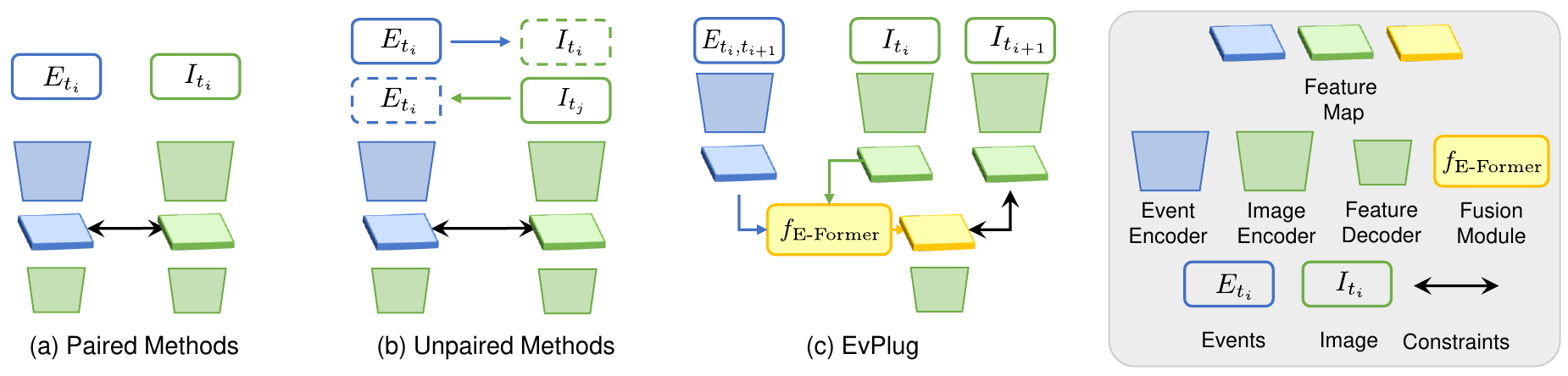}
  \caption{Illustration of differences among (a) paired methods~\cite{hu20nga, deng21learning}, (b) unpaired methods~\cite{gehrig20v2e, rebecq21e2vid, zhu21eventgan, wang21evdistill, wang21dual, messikommer22evtransfer}, and (c) \methodname{}. \methodname{} can learn event and image fusion connected by the event generation model. The legends on the right remain consistent throughout the paper.}
  \label{fig: method comparison}
  \vspace{-0.5cm}
\end{figure}

\subsection{Domain Adaption in Event-based Vision}

As both event streams and RGB images are visual signals, existing domain adaptation methods primarily exploit the consistency in geometry and texture between the two modalities to transfer knowledge from the RGB domain to event-based vision tasks. In terms of the data utilized, current methods can be divided into paired and unpaired approaches as shown in \Fref{fig: method comparison}. \textbf{Paired methods}~\cite{hu20nga, deng21learning} require pixel-wise alignment to ensure feature-level constraints, but this imposes higher demands on data acquisition. Although the DAVIS camera~\cite{licht08davis} can simultaneously output event streams and active pixel sensor (APS) frames, the APS frames lack color information and are of low quality.

\begin{table}[t]
    \centering
    \caption{Comparison about training and evaluation factors among \methodname{} and other methods.}
    \resizebox{\linewidth}{!}{
    \begin{tabular}{c|cc|ccccc}
    \toprule
    \multirow{3}{*}{Methods} & \multicolumn{2}{c}{Training} & \multicolumn{5}{c}{Evaluation} \\
    \cmidrule{2-8}
     & Model Input  & Supervision & \makecell[c]{Motion \\Blur} & HDR & \makecell[c]{Illumination \\Change} & \makecell[c]{Relatively\\ Static} & \makecell[c]{Temporal\\ Resolution}\\
     \hline
     RGB-based & Image & GT & & & \color{black}\Checkmark & \color{black}\Checkmark & Low\\
    Event-based & Events & GT & \color{black}\Checkmark & \color{black}\Checkmark &  &  & High\\
    Paired~\cite{deng21learning, hu20nga} & Pixel-wised Paired & RGB Model & \color{black}\Checkmark & \color{black}\Checkmark &  &  & High\\  
    \makecell[c]{Unpaired\\~\cite{wang21dual, messikommer22evtransfer, rebecq21e2vid, gehrig20v2e, wang21evdistill, zhu21eventgan}} & Unpaired & RGB Model & \color{black}\Checkmark & \color{black}\Checkmark &  &  & High\\  
    \hline
    \methodname{} & Paired & RGB Model & \color{black}\Checkmark & \color{black}\Checkmark & \color{black}\Checkmark & \color{black}\Checkmark & High\\  
    \bottomrule
    \end{tabular}
    }
    \label{tab: method comparison}
    \vspace{-0.5cm}
\end{table}

The core idea of \textbf{unpaired methods}~\cite{gehrig20v2e, rebecq21e2vid, zhu21eventgan, wang21evdistill, wang21dual, messikommer22evtransfer} is to convert between the two modalities to create paired data and then perform domain adaptation using the paired approach. However, since event streams and RGB images contain unmatching information (events lack absolute brightness values and color information, while images lack motion and illumination change information), this conversion process is ill-conditioned. Although EvTransfer~\cite{messikommer22evtransfer} has attempted to decouple the corresponding information, it does not fundamentally address the problem.
As summarized in \Tref{tab: method comparison}, these existing methods can only distill pure event-based models without multi-modal fusion.

\section{Method}
\label{method}

\begin{figure}[b]
    \vspace{-0.3cm}
  \centering

 \begin{minipage}[t]{0.48\textwidth}
    \includegraphics[width=\textwidth]{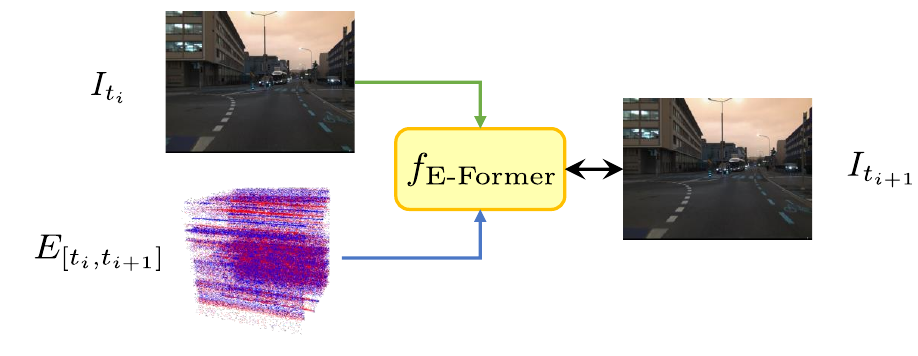}
    \caption{Feature-level constraint}
    \label{fig: feature level constraint}
 \end{minipage}
    \hfill
   \begin{minipage}[t]{0.48\textwidth}
    \includegraphics[width=\textwidth]{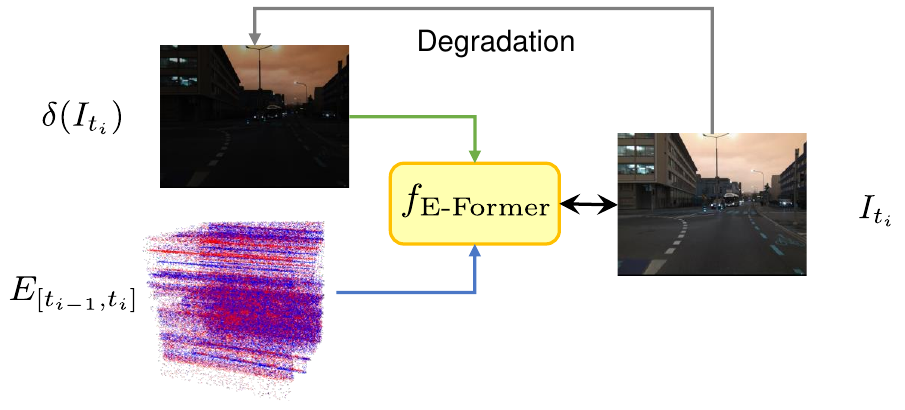}
    \caption{Event-image quality consistency}
    \label{fig: quality consistency}
 \end{minipage}
\end{figure}

\subsection{Event-Image Connection Constrained via Event Generation Model}
\label{non-concurrent fusion}

As model architectures~\cite{he16resnet, doso21vit} and data scales have been advanced, researchers have successfully trained high-performing vision models on RGB images~\cite{radford21clip, liu21swintransformer}. In contrast, event-based vision is in the early stage of its development. As shown in \Fref{fig: method comparison} (a) and (b), visual similarity between event streams and RGB images encourages researchers to transfer knowledge from RGB images to events and such a process can be explained: 
\begin{equation}
    E_{[t_{i-1}, t_i]} \leftrightarrow I_{t_i},
    \label{equation paired constraint}
\end{equation}
where $E_{[t_{i-1}, t_i]}$ represents event streams from time $t_{i-1}$ to $t_i$, $I_{t_i}$ is the RGB image at time $t_i$, $\leftrightarrow$ means the information connection between images and events, such as feature map constraint~\cite{wanglin22kd_survey}.

However, such a straightforward mapping method does not work well due to the different sensing in motion, texture, and color between events and RGB images. 
This issue is caused by the imaging mechanism of event cameras, a.k.a. event generation model~\cite{hidalgo22edso}.

Event cameras~\cite{licht08davis} generate asynchronous event streams by measuring per-pixel brightness changes.
An event $e_i = (x_i, y_i, t_i, p_i)$ occurs at pixel $(x_i, y_i)$ at time $t_i$ when the logarithmic brightness change reaches the threshold:
\begin{equation}
    \log I(x_i, y_i, t_i) - \log I(x_i, y_i, t_p) = p_i \cdot C,
    \label{event generation model}
\end{equation}
where $t_p$ is the timestamp of the last event at pixel $(x_i, y_i)$, $p_i \in \{ -1, 1\}$ is the polarity, $C$ is the threshold.
Previous research on video frame interpolation and deblurring \cite{tulyakov22timelens++, han21evintsr, teng22nest} have shown that, based on an image at time $t_i$ and the event stream following $t_i$, a frame at any future time can be reconstructed.
When images and events are pixel-wisely aligned, such relation is denoted as:
\begin{equation}
    I(x_{i+1}, y_{i+1}, t_{i+1}) = I(x_i, y_i, t_i) \cdot \text{exp}(\sum_{e_j \in E_{[t_i, t_{i+1}]}}{p_j \cdot C}).
\end{equation}

Hence, we propose the following information constraint connecting event streams and RGB images:
\begin{equation}
    I_{t_i} + E_{[t_i, t_{i+1}]} \leftrightarrow I_{t_{i+1}},
    \label{equation evplug constraint}
\end{equation}
where "$+$" refers to a multi-modal data fusion process, as shown in \Fref{fig: method comparison} (c).

In addition to expressing the physical constraints between event cameras and RGB cameras according to their image formation model, this constraint also implies that it is possible to predict the results at any future moment using a single RGB image and subsequent event streams.

\subsection{Feature-level Multi-modal Fusion}
\label{rgb-event fusion}

When fusing multi-modal data, the feature-level constraint allows effective information fusion for complementary use of both modalities of data~\cite{daniel21ramnet, zhou2023rgb, wanglin22kd_survey}. 
This inspires us to extend the information constraint in \Eref{equation evplug constraint} to feature level as illustrated in \Fref{fig: feature level constraint}:
\begin{equation}
    f_{\text{E-Former}}(f_{\text{ImEncoder}}(I_{t_i}), f_{\text{EvEncoder}}(E_{[t_i, t_{i+1}]})) \leftrightarrow f_{\text{ImEncoder}}(I_{t_{i+1}}),
    \label{feature level constraint}
\end{equation}
where $f_{\text{E-Former}}$ is the fusion module, $f_{\text{ImEncoder}}$ and $f_{\text{EvEncoder}}$ are the image and the event encoders.

We design $f_{\text{E-Former}}$ based on the transformer decoder layer~\cite{vaswani17transformer}, which can perform feature association and does not require strict pixel-wise alignment between event streams and RGB images, alleviating the burden of data acquisition. 
Within the decoder layers, event features serve as the input for cross-attention keys and values, while the image features are updated through multiple decoder layers to obtain the fused features.
In terms of information constraints, the feature-level constraint enables $f_{\text{E-Former}}$ to learn rich information from RGB-based models~\cite{wanglin22kd_survey}. Besides, this constraint does not change the structure and weights of the RGB-based model, and $f_{\text{EvEncoder}}$ and $f_{\text{E-Former}}$ can be plug-and-play for evaluation.

\begin{figure}[t]
  \centering
   \includegraphics[width=\textwidth]{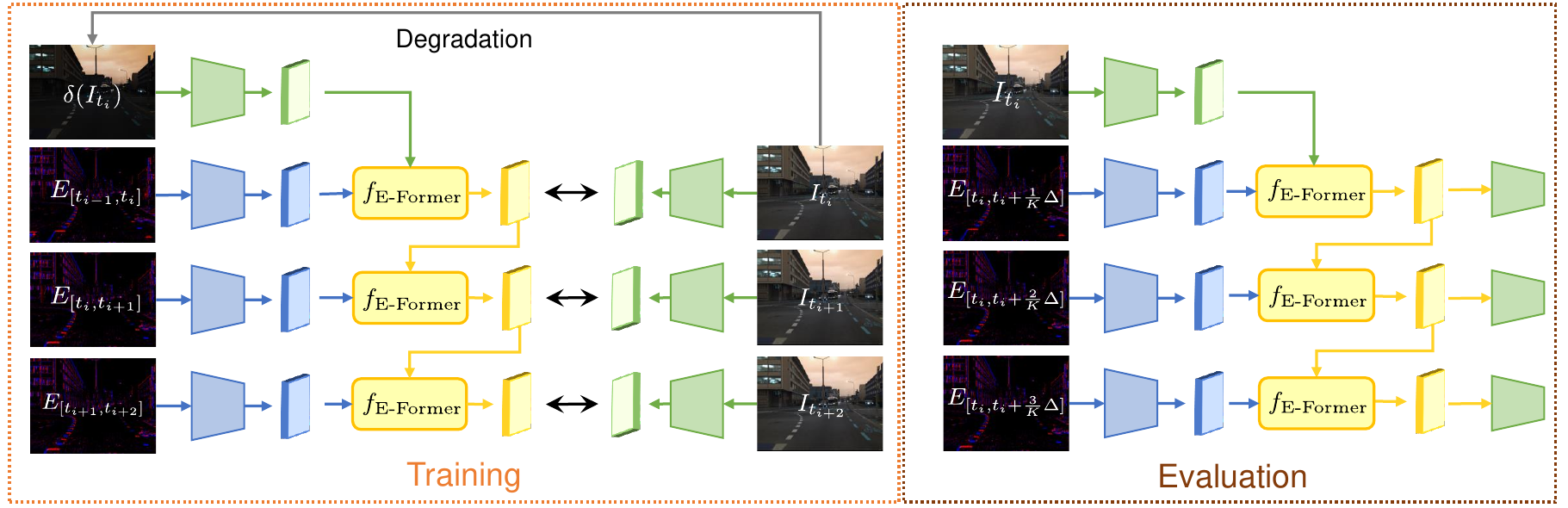}
  \caption{Illustration of \methodname{} pipeline. During training, event encoder $f_{\text{EvEncoder}}$ (blue) and fusion module $f_{\text{E-Former}}$ (yellow) are learned to ensure the event-image quality consistency and temporal consistency under the supervision of RGB sequential features from the fixed RGB-based model (green).
  During the evaluation, $f_{\text{EvEncoder}}$ and $f_{\text{E-Former}}$ can work as plug-and-play modules for complementary use of both modalities for robust inference with high temporal resolution.
  }
  \label{fig: pipeline}
  \vspace{-0.4cm}
\end{figure}

However, using the constraint in \Eref{feature level constraint} to train $f_{\text{E-Former}}$ faces two challenges:
1) It cannot utilize event streams $E_{[t_{i-1}, t_i]}$ for robust inference on image $I_{t_i}$;
2) it cannot assure temporal consistency when performing model inferences with high temporal resolution using events between adjacent RGB images. 
To tackle these challenges, we propose two constraints to enable fusion module $f_{\text{E-Former}}$ to learn event-image quality consistency and temporal consistency.

\paragraph{Event-Image Quality Consistency}
Existing RGB-based vision models~\cite{doso21vit, he16resnet} usually extract high-quality features from normal-quality RGB images. 
Due to the limited imaging characteristics of RGB cameras in HDR or fast motion scenes, the resulting images may suffer from degradation, thus causing distortion in the feature space. 
So one of the goals of data fusion module $f_{\text{E-Former}}$ is to rectify the feature space distortion caused by RGB image degradation with event streams.

For training, it is hard to obtain paired normal and degraded RGB sequences of the same view in the real world.
To tackle this issue, we add photometric degradation on normal-quality RGB sequences to get synthetic paired degraded images.
As shown in \Fref{fig: quality consistency}, this can be formulated as:
\begin{equation}
    f_{\text{E-Former}}(f_{\text{ImEncoder}}(\delta(I_{t_i})), f_{\text{EvEncoder}}(E_{[t_{i-1}, t_i]})) \leftrightarrow f_{\text{ImEncoder}}(I_{t_i}),
    \label{degradation constraint}
\end{equation}
where $\delta(\cdot)$ is the image degradation function.
For applying synthetic degradation, we introduce two types of operations: strong/low light with brightness contrast augmentation in PyTorch~\cite{paszke19pytorch} and motion blur with pre-computed optical flow.
We warp the original image with optical flow following~\cite{farneback03flow} in OpenCV to interpolate frames and average them for simulating motion blur.

\paragraph{Event-Image Temporal Consistency}
Directly using the constraint in \Eref{feature level constraint} for inference with high temporal resolution cannot assure the temporal consistency of inference results.
After obtaining fusion features at time $t_i$ as:
\begin{equation}
    F_{t_i} = f_{\text{E-Former}}(f_{\text{ImEncoder}}(\delta(I_{t_i})), f_{\text{EvEncoder}}(E_{[t_{i-1}, t_i]})),
\end{equation}
we use $f_{\text{E-Former}}$ to iteratively update the fusion feature with events to assure temporal consistency:
\begin{equation}
    F_{t_j} = f_{\text{E-Former}}(F_{t_{j-1}}, f_{\text{EvEncoder}}(E_{[t_{j-1}, t_j]})), \quad j=i+1, i+2, ..., i+K,
    \label{eq: iteratively update}
\end{equation}
where $F_{t_j}$ is the fusion feature at time $t_j$, and $K$ is the fusion step.

Thus we get sequential constraints to enable $f_{\text{E-Former}}$ to learn temporal consistency:
\begin{equation}
    F_{t_j} \leftrightarrow f_{\text{ImEncoder}}(I_{t_j}), \quad j=i+1, i+2, ..., i+K.
    \label{eq: seq constraint}
\end{equation}

By integrating constraints from ~\Eref{degradation constraint} and \Eref{eq: seq constraint}, $f_{\text{E-Former}}$ can utilize events $E_{[t_{i-1}, t_i]}$ to rectify the degradation of the image $I_{t_i}$ and assure the temporal consistency of evaluation results with subsequent events in an iterative way.

\subsection{Training and Evaluation}
\paragraph{Training}
Our training procedure is illustrated in \Fref{fig: pipeline} (left).
As mentioned in \Eref{feature level constraint}, feature $F_{t_j}$ should be consistent with image feature $f_{\text{ImEncoder}}(I_{t_j})$.
In our work, we use two losses to measure the constraints: task-level loss $\loss_{\text{task}}$ and feature-level losses.
Task-level loss is defined by the task-specific training loss under supervision from outputs of RGB-based model on original RGB image inputs.
Following \cite{hu20nga}, feature-level losses consist of feature reconstruction loss $\loss_{\text{recon}}$ and feature style loss $\loss_{\text{style}}$:
\begin{eqnarray}
    && \loss_{\text{recon}}(t_j) = \text{MSE}(F_{t_j}, f_{\text{ImEncoder}}(I_{t_j})), \\
    && \loss_{\text{style}}(t_j) = \text{MSE}(\text{Gram}(F_{t_j}), \text{Gram}(f_{\text{ImEncoder}}(I_{t_j}))),
\end{eqnarray}
where $\text{MSE}$ is mean squared error, $\text{Gram}$ is the Gram matrix~\cite{gatys16imagestyle}.

Thus the final loss becomes:
\begin{equation}
    \loss_{\text{all}} = \sum_{j}{\lambda_{\text{task}} \loss_{\text{task}}(t_j) +  \lambda_{\text{recon}} \loss_{\text{recon}}(t_j) + \lambda_{\text{style}} \loss_{\text{style}}(t_j)}, \quad j=i, i+1, ..., K,
\end{equation}
where $\lambda_{\text{task}}$, $\lambda_{\text{recon}}$, and $\lambda_{\text{style}}$ are loss weights.

\paragraph{Evaluation}
As shown in \Fref{fig: pipeline} (right), during the evaluation stage, robust features are first obtained by fusing the image at time $I_{t_i}$ and the event stream $E_{[t_{i-1}, t_i]}$, which are used for subsequent decoding to obtain robust inference results. 
Next, the incoming high temporal resolution event stream is divided into $K$ event slices with equal time interval (assuming the duration of neighboring RGB images is $\Delta$). 
These slices are then sequentially iterated and updated through the fusion module, and the output features are obtained through the feature decoder module. 
Through this iterative process, \methodname{} has $K$ times temporal resolution than original RGB-based methods.
Since $K$ can be adjusted according to the different applications, \methodname{} allows for flexible high-temporal-resolution inference.

\section{Experiments}

\paragraph{Baselines}
For comparing with domain adaptation methods, we choose NGA~\cite{hu20nga} as the pixel-aligned paired method, and for unpaired methods, we select E2VID~\cite{rebecq21e2vid} and EV-Transfer~\cite{messikommer22evtransfer}, which have released open-source codes.
V2E~\cite{gehrig20v2e}, works as an event simulator, inevitably faces sim-to-real gap compared with paired method NGA~\cite{hu20nga}, and we don't evaluate here.
Although EvDistill~\cite{wang21evdistill} and DTL~\cite{wang21dual} are unpaired methods, but there is no public training code to employ them on our datasets.
To validate the benefits of fusing the two modalities, we compare supervised RGB-based methods for each task. 

\paragraph{Downstream Tasks}
The middle- and high-level vision datasets that contain both RGB images and event streams are generally scarce.
Considering the data availability, we evaluate the performance of \methodname{} on three downstream vision tasks: object detection on DSEC-MOD~\cite{zhou2023rgb}, semantic segmentation on DSEC-Semantic~\cite{sun22ess}, and 3D hand pose estimation on EvRealHands~\cite{jiang23evhandpose}. 
We highly recommend readers to see the results of high temporal resolution in the supplementary video.

\paragraph{Implementation Details of \methodname{}}
\methodname{} has a simple and generalized architecture.
In all tasks, event streams are represented as voxels~\cite{zhu19evflow, victor21eventhands}, and the event encoder $f_{\text{EvEncoder}}$ shares the same architecture with the image encoder $f_{\text{ImEncoder}}$ except for the channel dimension of the first CNN layer.
The fusion module $f_{\text{E-Former}}$ is composed of $N$ sequential transformer decoder layers~\cite{vaswani17transformer} ($N$ is 3 in our experiments).
In training, sequential fusion step $K$ is set as 2, $\lambda_{\text{recon}}$ as 10, $\lambda_{\text{style}}$ as 10.
We use Adam~\cite{adam15} with learning rate 0.001 to optimize the $f_{\text{EvEncoder}}$ and $f_{\text{E-Former}}$.
Apart from these settings, data processing and technical details are consistent with the RGB-based methods.
In our experiments, \methodname{} is trained on a single TITAN RTX in two days.

\subsection{Object Detection}
\paragraph{Experimental Setup}
We validate \methodname{} on the DSEC-MOD dataset~\cite{zhou2023rgb} for moving object detection with 10495 frames for training and 2819 frames for evaluation.
For the RGB-based backbone of knowledge distillation methods and RGB-based methods, we use DETR~\cite{carion20detr}, the widely used backbone for object detection, as the trade-off between accuracy and efficiency.
We use the pretrained DETR-R50 model on COCO~\cite{lin14coco} and finetune it on DSEC-MOD~\cite{zhou2023rgb} for 10 epochs with batchsize 64 to get an RGB-based model.
For the event-based method, we represent events as voxels as \cite{zhu19evflow} and use the same backbone as DETR-R50, which requires training for 30 epochs.
For the RGB-event fusion method, we use the released model of RENet~\cite{zhou2023rgb} for evaluation.
In order to uniformly evaluate the various methods using the same standard, we use the detection metrics of COCO~\cite{lin14coco} (reported as \%, $\uparrow$ means higher is better, while $\downarrow$ is the opposite).

\begin{figure}[b]
  \centering
    \vspace{-0.1cm}
  \includegraphics[width=\textwidth]{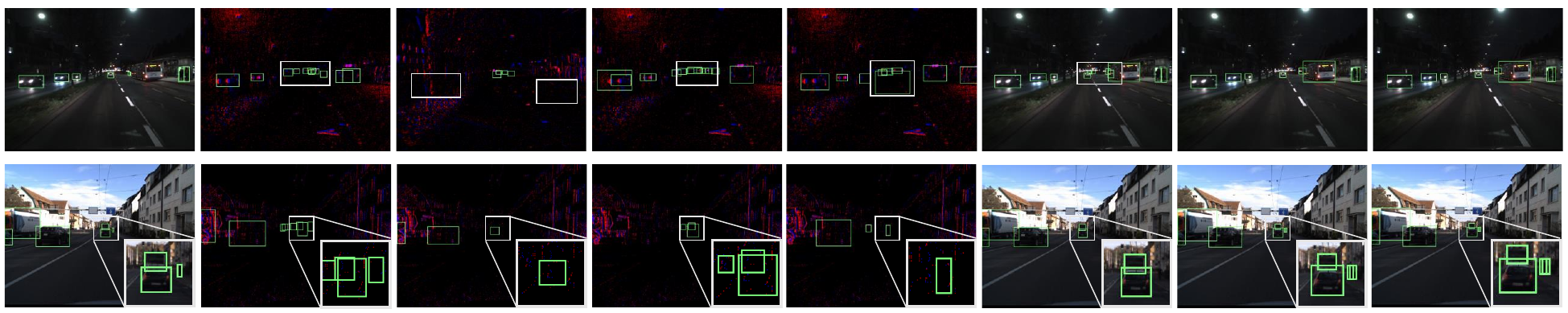}
  \begin{flushleft}
    \scriptsize  \hspace{0.4cm}RGB(GT) \hspace{0.4cm} Event-based~\cite{carion20detr} \hspace{0.4cm} NGA~\cite{hu20nga} \hspace{0.5cm} E2VID~\cite{rebecq21e2vid} \hspace{0.2cm} EvTransfer~\cite{messikommer22evtransfer}  \hspace{0.1cm} RGB-based~\cite{carion20detr}\hspace{0.15cm} \methodname{} (w/o $\delta$) \hspace{0.4cm} \methodname{}
  \end{flushleft}
  \caption{Qualitative results on DSEC-MOD~\cite{zhou2023rgb}. Green boxes are bounding boxes and gray boxes are spaces zoomed in.}
  \label{fig: quality od}
  \vspace{-0.5cm}
\end{figure}

\paragraph{Results}
As quantitative results shown in \Tref{tab: object detection results}, \methodname{} outperforms the domain adaptation methods~\cite{hu20nga, rebecq21e2vid, messikommer22evtransfer} by more than 20 $\text{AP}_{\text{50}}$, which can be attributed to its ability to fuse event streams with RGB images, rather than relying solely on events for inference.
Compared with RENet~\cite{zhou2023rgb} which designs specific architecture for object detection and requires training from scratch, \methodname{} can utilize pretrained models from large-scale image dataset~\cite{lin14coco}, thus resulting in 18.2 $\text{AP}_{\text{50}}$ improvement.
\methodname{} brings only 2.0 $\text{AP}_{\text{50}}$ improvement for the RGB-based model, because DSEC-MOD~\cite{zhou2023rgb} contains few data under challenging scenes (HDR, fast motion) where events have better imaging quality over RGB images.
Qualitative results in \Fref{fig: quality od} show that \methodname{} can make the RGB-based model sensitive to small moving objects in the distance, which is due to the rich motion information contained in event streams.



\begin{table}[t]
    \centering
    \vspace{-0.5cm}
    \caption{Quantitative results on object detection and semantic segmentation.}
    \resizebox{\linewidth}{!}{
    \begin{tabular}{c|c|c|ccccc}
    \toprule
    \multirow{2}{*}{Methods} & \multirow{2}{*}{Input} & \multirow{2}{*}{Test Time} & \multicolumn{3}{c}{Object Detection} & \multicolumn{2}{c}{\makecell[c]{Semantic Segmentation}}\\
    & & &  AP $\uparrow$ & $\text{AP}_{\text{50}}$ $\uparrow$ & $\text{AP}_{\text{75}}$ $\uparrow$ & mIoU $\uparrow$ & Acc $\uparrow$\\
    \hline
    Event-based & $E_{[t_{i-1}, t_i]}$ & $t_i$ &18.5 & 31.4 & 14.1 & 51.2 & 87.7 \\
    NGA~\cite{hu20nga}& $E_{[t_{i-1}, t_i]}$ & $t_i$& 19.0 & 32.2 & 15.4 & 50.8 & 87.4 \\
    E2VID~\cite{rebecq21e2vid} & $E_{[t_{i-1}, t_i]}$& $t_i$ & 15.3 & 24.5 & 13.6 & 38.7 & 74.6 \\
    EvTransfer~\cite{messikommer22evtransfer} & $E_{[t_{i-1}, t_i]}$& $t_i$ & 17.6 & 27.4 & 14.4 & 49.9& 87.6 \\
    \hline
    ESS~\cite{sun22ess}& $I_{t_i}$, $E_{[t_{i-1}, t_i]}$& $t_i$ & -- & --  & --  & 53.3 & 89.4 \\
    RENet~\cite{zhou2023rgb} & $I_{t_i}$, $E_{[t_{i-1}, t_i]}$& $t_i$ & 18.5 & 36.9 & 17.2 & -- & -- \\
     RGB-based & $I_{t_i}$ & $t_i$& 29.6 & 53.1 & 28.8 & 63.8 & 93.2\\
   \hline
    \methodname{} & $I_{t_i}$,$E_{[t_{i-1}, t_i]}$ & $t_i$ & 30.1 (\textbf{0.5$\uparrow$}) & 55.1 (\textbf{2.0$\uparrow$}) & 30.1 (\textbf{1.3$\uparrow$}) & 63.2 & 93.1\\
    \hline
    \methodname{} (w/o $\delta$) & $I_{t_i}$, $E_{[t_{i-1}, t_i]}$ & $t_i$ & 29.6 & 54.3 & 28.8 & 63.7 & 93.2 \\
    \methodname{}  & $I_{t_i}$, $E_{[t_{i-1}, t_{i+2}]}$ & $t_{i+2}$ & 19.0 & 40.6 & 16.4 & 53.5 & 90.7 \\
    \methodname{} (w/o iter) & $I_{t_i}$, $E_{[t_{i-1}, t_{i+2}]}$ & $t_{i+2}$  & 18.7 & 40.2 & 16.1 & 52.4 & 90.1 \\
    \bottomrule
    \end{tabular}
    }
    \vspace{-0.3cm}
    \label{tab: object detection results}
\end{table}

\subsection{Semantic Segmentation}
\paragraph{Experimental Setup}
We validate \methodname{} on DSEC-Semantic~\cite{sun22ess} (8082 frames for training, 2809 frames for evaluation) for semantic segmentation.
For the RGB-based model, we use the extension version of DETR~\cite{carion20detr} on panoptic segmentation~\cite{Kirillov19panoptic} and finetune it on Cityscapes~\cite{Cordts2016Cityscapes}.
For the event-based method, we use supervised Ev-SegNet~\cite{alonso19evsegnet} for comparison.
Apart from the mentioned domain adaptation methods, ESS~\cite{sun22ess} proposed an unsupervised domain adaptation method.
We use the same evaluation metric (mIoU (\%) and Accuracy (Acc., \%)) as ESS~\cite{sun22ess} for a fair comparison.

\begin{figure}[t]
  \centering
  \includegraphics[width=\textwidth]{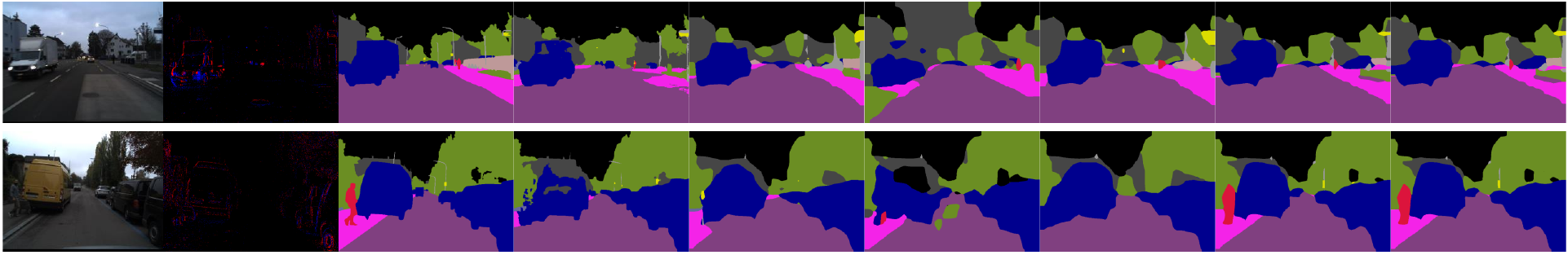}
  \begin{flushleft}
\scriptsize  \hspace{0.5cm}RGB \hspace{0.8cm} Events \hspace{1cm} GT \hspace{0.55cm}Ev-SegNet~\cite{alonso19evsegnet}\hspace{0.4cm}NGA~\cite{hu20nga} \hspace{0.3cm} E2VID~\cite{rebecq21e2vid} \hspace{0.1cm} EvTransfer~\cite{messikommer22evtransfer}  \hspace{0.15cm} DETR~\cite{carion20detr} \hspace{0.55cm} \methodname{}
  \end{flushleft}
  \caption{Qualitative results on DSEC-Semantic~\cite{sun22ess}.}
  \label{fig: quality ss}
  \vspace{-0.2cm}
\end{figure}

\paragraph{Results}
Results in \Tref{tab: object detection results} and \Fref{fig: quality ss} show that \methodname{} outperforms mentioned domain adaptation methods~\cite{hu20nga, rebecq21e2vid, sun22ess, messikommer22evtransfer} due to the multi-modal fusion of both modalities.
However, \methodname{} achieves a bit lower accuracy compared with the RGB-based method.
The reason is that color and texture play a significant role in semantic segmentation, and there are few degraded RGB images in the DSEC-Semantic~\cite{sun22ess}. Compared to the gain that events bring to RGB images, the disruption caused by event features to the RGB feature space has a greater impact.
Results in \Tref{fig: quality od} (methods: \methodname{}, test time: $t_{i+2}$) show that when performing semantic segmentation with high temporal resolution, \methodname{} outperforms event-based methods~\cite{alonso19evsegnet} by over 2.5 \% mIoU.
Due to the complementary nature of RGB images at time $t_i$ to event streams $E_{[t_{i}, t_{i+2}]}$, \methodname{} achieves higher accuracy than methods based solely on event streams.
This implies that \methodname{}, at the cost of sacrificing a small portion of accuracy on RGB images, enables the RGB-based model with high temporal resolution inference.

\subsection{Hand Pose Estimation}

\begin{table}[b]
    \centering
    \vspace{-0.4cm}
    \caption{Quantitative results on 3D hand pose estimation.}
    \resizebox{\linewidth}{!}{
    \begin{tabular}{c|c|c|cccccc}
    \toprule

    \multirow{2}{*}{Methods}& \multirow{2}{*}{Input} & \multirow{2}{*}{Test Time} & \multicolumn{2}{c}{Normal} & \multicolumn{2}{c}{Strong Light}  & \multicolumn{2}{c}{Flash}\\
    & & & MPJPE $\downarrow$ & AUC $\uparrow$ & MPJPE $\downarrow$ & AUC $\uparrow$ & MPJPE $\downarrow$ & AUC $\uparrow$\\
    \hline
    Event-based~\cite{victor21eventhands}& $E_{[t_{i-1}, t_i]}$ & $t_i$ & 29.16  & 77.7 & 32.48 & 71.7 & 52.92 & 60.9\\
    NGA~\cite{hu20nga}& $E_{[t_{i-1}, t_i]}$ & $t_i$ & 20.48 & 79.6 & 28.37 & 71.8 & 41.06  & 61.4\\
    E2VID~\cite{rebecq21e2vid}& $E_{[t_{i-1}, t_i]}$ & $t_i$ & 29.81 & 73.6 & 30.77 & 70.3 & 42.35 & 61.0\\
    EvTransfer~\cite{messikommer22evtransfer} & $E_{[t_{i-1}, t_i]}$ & $t_i$ & 22.67 & 77.5 & 29.39 & 70.8 & 41.32 & 60.6 \\
    \hline
    RGB-based~\cite{cho22fastmetro}& $I_{t_i}$ & $t_i$  & 14.31  & 85.6 & 42.43  & 59.1 & 26.31 & 73.9\\
    \hline
    \methodname{} & $I_{t_i}$,$E_{[t_{i-1}, t_i]}$ & $t_i$  & 13.78 (\textbf{0.53$\downarrow$}) & 86.2 (\textbf{0.6$\uparrow$}) & 28.16 (\textbf{14.27$\downarrow$}) & 71.9 (\textbf{12.8$\uparrow$}) & 26.00 (\textbf{0.31$\downarrow$}) & 74.3 (\textbf{0.4$\uparrow$})\\
    \hline
    \methodname{} (w/o $\delta$) & $I_{t_i}$,$E_{[t_{i-1}, t_i]}$ & $t_i$  & 13.78  & \textbf{86.2} & 46.07 &  56.0 & 26.85  & 73.4 \\
    \methodname{}  & $I_{t_i}$, $E_{[t_{i-1}, t_{i+2}]}$ & $t_{i+2}$ & 18.28 & 81.8 & 30.10 &  70.8& 35.34 & 67.3 \\
    \methodname{} (w/o iter) & $I_{t_i}$, $E_{[t_{i-1}, t_{i+2}]}$ & $t_{i+2}$ & 18.39 & 81.6 & 30.25 & 70.0  & 36.12 & 65.4 \\
    \bottomrule
    \end{tabular}
    }
    \vspace{-0.4cm}

    \label{tab: hand pose estimation results}
\end{table}

\paragraph{Experimental Setup}
We evaluate the performance of \methodname{} on hand pose estimation on dataset EvRealHands~\cite{jiang23evhandpose}, a multi-modal hand dataset consisting of 4452 seconds of event streams from DAVIS346 (346$\times$260 pixels) and corresponding RGB sequences from FLIR cameras (2660$\times$2300 pixels).
We perform comparisons on the sequences under different illumination and hand movements: normal, strong light, flash, and fast motion.
For RGB-based baseline and domain adaptation backbones, we select FastMETRO~\cite{cho22fastmetro} which performs high accuracy with low computational cost.
For efficiency, we use the lightweight version (a ResNet34~\cite{he16resnet} backbone and 4 transformer~\cite{vaswani17transformer} layers).
For event-based methods, we use EventHands~\cite{victor21eventhands}.
We follow the implementation details of EventHands~\cite{victor21eventhands} and FastMETRO~\cite{cho22fastmetro} to pre-process event streams and RGB images.
In our setup, RGB images are cropped in 192$\times$192 and event streams in 128$\times$128.
We use the common MPJPE (root-aligned mean per joint position error in Euclidean distance (mm)) and AUC (reported as \%) metrics for quantitative evaluation.

\paragraph{Results}

\begin{figure}[t]
  \vspace{-0.5cm}
  \centering
  \includegraphics[width=\textwidth]{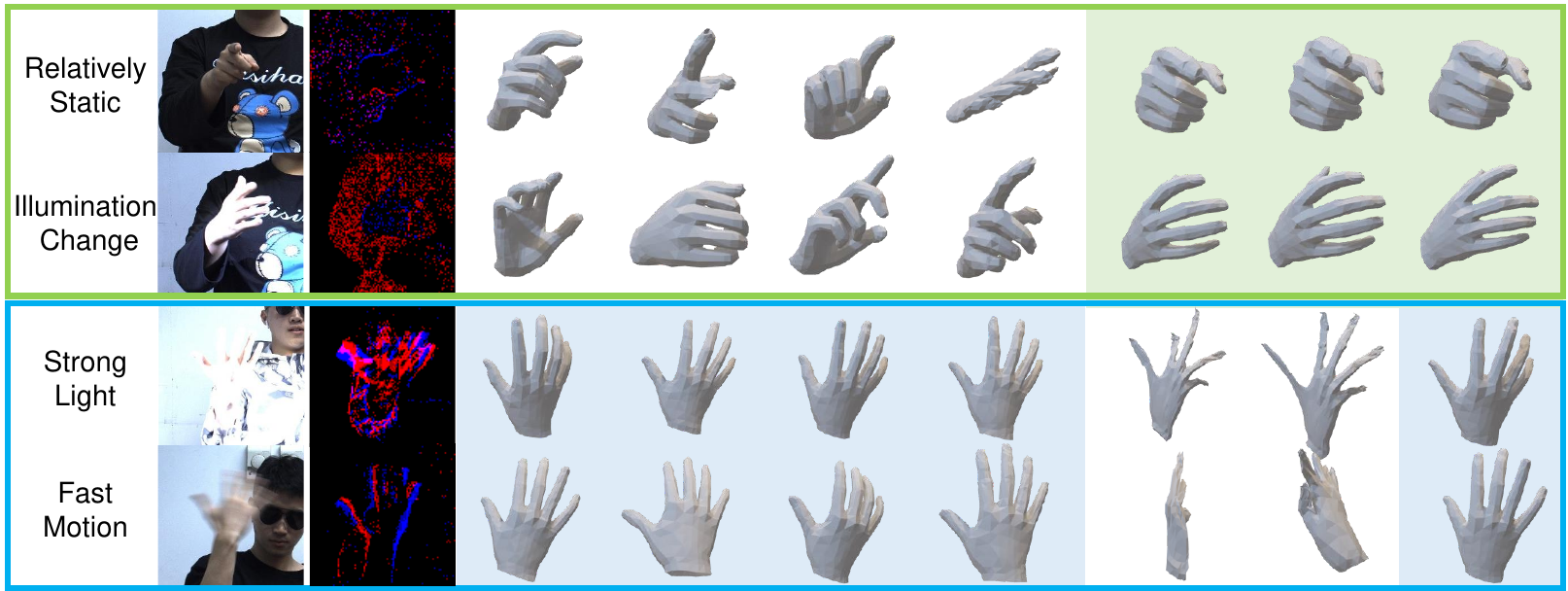}
      \begin{flushleft}
    \tiny  \hspace{1.9cm}RGB \hspace{0.75cm} Events \hspace{0.25cm} EventHands~\cite{victor21eventhands} \hspace{0.15cm}  NGA~\cite{hu20nga} \hspace{0.25cm} E2VID~\cite{rebecq21e2vid} \hspace{0.25cm} EvTransfer~\cite{messikommer22evtransfer} \hspace{0.cm} FastMETRO~\cite{cho22fastmetro} \hspace{0.0cm} \methodname{} (w/o $\delta$) \hspace{0.15cm} \methodname{}
  \end{flushleft}
   \caption{Qualitative results on EvRealHands~\cite{jiang23evhandpose}.}
  \label{fig: quality hpe}
  \vspace{-0.2cm}
\end{figure}

As quantitative results are shown in \Tref{tab: hand pose estimation results}, 
\methodname{} outperforms domain adaptation methods~\cite{hu20nga, rebecq21e2vid, messikommer22evtransfer} over 6 mm MPJPE lower in normal scenes and 15 mm MPJPE on flash scenes. This benefits from the fact that \methodname{} is an event and image fusion method, rather than a pure event-based domain adaptation method.
The plug-and-play module learned by our method will significantly improve the robustness of the RGB-based model in strong light scenes, resulting in a 14.27 mm lower MPJPE in such conditions.
This demonstrates that our method can integrate the HDR robustness characteristics of event cameras into the RGB-based model.
Qualitative results in \Fref{fig: quality hpe} show that these domain adaptation methods~\cite{hu20nga, rebecq21e2vid, messikommer22evtransfer} cannot handle the scenes when hands are relatively static or the illumination changes. And RGB-based models cannot achieve robust hand pose estimation when images degrade, such as overexposure and motion blur.
\methodname{} can perform robust hand pose estimation in these challenging issues, which derives from the complementary use of two modalities.

\subsection{Ablation Study}

In the ablation study, we demonstrate the effectiveness of the two constraints for image-event quality consistency and temporal consistency.

\paragraph{Event-Image Quality Consistency}
To validate the effectiveness of the quality constraint, we can compare the performance of \methodname{} at time $t_i$ with and without incorporating $\delta$ in \Eref{degradation constraint} (denoted as "w/o $\delta$") in the training procedure.
Quantitative results in \Tref{tab: object detection results} and \Tref{tab: hand pose estimation results} show that this constraint can improve the performance of \methodname{} on object detection (DSEC-MOD~\cite{zhou2023rgb}) and 3D hand pose estimation (EvRealHands~\cite{jiang23evhandpose}), especially under strong lights (18.09 mm lower on MPJPE ).
However, it will result in a small decrease in metrics on semantic segmentation, which mainly results from the absence of degraded images in evaluation data.
Qualitative results in \Fref{fig: quality od} and \Fref{fig: quality hpe} show that this constraint can improve the performance of \methodname{} under HDR imaging and fast motion scenes.

\paragraph{Event-Image Temporal Consistency}
A direct approach (denoted as "w/o iter") to validate the effectiveness of the temporal consistency constraint is to integrate the features $F_{t_i}$ at time $t_i$ with events during evaluation:
\begin{equation}
    F_{t_j} = f_{\text{E-Former}}(F_{t_{i}}, f_{\text{EvEncoder}}(E_{[t_{j-1}, t_j]})), \quad j=i+1, i+2, ..., i+K,
\end{equation}
which is different from \Eref{eq: iteratively update}.
Therefore we can compare the results at time $t_{i+2}$ with the image $I_{t_i}$ and the event stream $E_{[t_{i-1}, t_i]}$ to validate the efficiency of event-image temporal consistency.
Quantitative results in \Tref{tab: object detection results} and \Tref{tab: hand pose estimation results} show that this constraint will contribute to a 0.3 lift on AP on object detection, 1.1 \% lift on mIoU on semantic segmentation, and 0.1 mm MPJPE improvement on hand pose estimation.

\section{Conclusion}
In this paper, we propose a unified framework called \methodname{} that learns a plug-and-play event and image fusion module under supervision from the RGB-based model connected by the event generation model. Our method only requires unlabeled, non-strictly pixel-aligned image-event data pairs for training. Working as a plug-in for the RGB-based model, \methodname{} enables the RGB-based model with robustness to HDR imaging and high-temporal-resolution inference. We demonstrate the effectiveness of \methodname{} on three vision tasks: object detection, semantic segmentation, and hand pose estimation.

\paragraph{Limitations}
In our fusion model $f_{\text{E-Former}}$, when the scale of the feature map is large, the computational cost of the transformer-based fusion module can be quite high. Additionally, although $f_{\text{E-Former}}$ successfully bridges the temporal consistency between neighboring images and events, it cannot model long-term consistency across multiple images. 
Therefore, designing a fusion framework that can model long-term temporal consistency with low computational cost is a promising research direction. 

{\small
\bibliographystyle{ieee_fullname}
\bibliography{egbib}
}

\end{document}